# Design for a Darwinian Brain: Part 2. Cognitive Architecture

March 2013

*Chrisantha Fernando, Vera Vasas*

Department of Electronic Engineering and Computer Science (EECS)*,* Queen Mary University of London, Mile End Road, London

**Abstract**

The accumulation of adaptations in an open-ended manner during lifetime learning is a holy grail in reinforcement learning, intrinsic motivation, artificial curiosity, and developmental robotics. We present a specification for a cognitive architecture that is capable of specifying an unlimited range of behaviors. We then give examples of how it can stochastically explore an interesting space of adjacent possible behaviors. There are two main novelties; the first is a proper definition of the fitness of self-generated games such that interesting games are expected to evolve. The second is a modular and evolvable behavior language that has systematicity, productivity, and compositionality, i.e. it is a physical symbol system. A part of the architecture has already been implemented on a humanoid robot.

**Introduction**

"The main objective of the developmental robotics field is to create "open-ended, autonomous learning systems that continually adapt to their environment" [12], as opposed to constructing robots that carry out particular, predefined tasks." (Bellas, Duro et al. 2010). We present a design for a cognitive architecture capable of ontogenetic evolution of open-ended controllers and open-ended fitness functions (i.e. games). When implemented in a robot, the architecture is intended to generate playful interaction with the world, in which intuitively 'interesting' behaviors are discovered. The architecture in a sense provides a computational definition of what 'interesting' means. Eventually it can reasonably be expected to exhibit creative and curious playful behavior when coupled with any situated and embodied device that has rich sensorimotor contingencies, i.e. where actions have sensory consequences.

The overall cognitive architecture consists of two evolving populations and a memory. The first population contains a set of actor molecules; the second population contains a set of game molecules, see Figure 1. The memory is of two types, a short-term working memory for data used by the molecules, and a long-term memory for the static storage of previously evolved useful molecules.

Living Machines 2013 Natural History Museum, London.

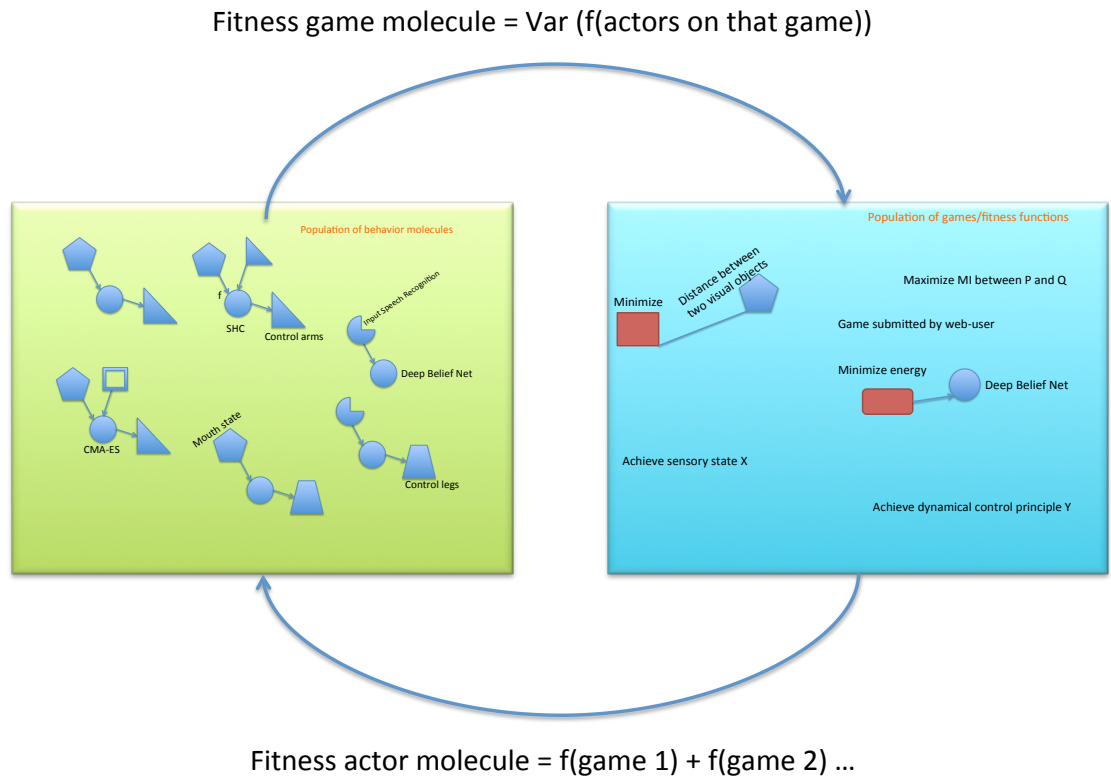

Figure 1. The overall Darwinian cognitive architecture. On the left there is a population of actor molecules, and on the right is a population of game molecules. The fitness of actor molecules is determined by the game molecules, and the fitness of game molecules is dependent on the fitness of actor molecules.

The fitness of molecules in one population depends upon the molecules in the other population. The fitness $f_{aj}$ of an actor molecule $j$ in the first population is the aggregate fitness of the component fitness $f_{aj,gi}$ achieved by the actor molecule $j$ on the $N$ game molecules in the second population, i.e. $f_{aj} = f_{aj,g1} + f_{aj,g2} + \ldots f_{aj,gN}$. The equation below gives the fitness of an actor $a$ on the $g$ games associated with it G.

$$F_a(G) = \sum_{g \in G_{assoc}(a)} F(a,g)$$

The fitness of a game molecule $i$, $f_{gi}$, in the second population is the variance of fitness components of the $M$ actor molecules in the first population on that game, i.e. $f_{gi} = Var(f_{a1,gi}, f_{a2,gi}, \ldots, f_{aM,gi})$. The equation below shows that the fitness of a game atom is the variance of the fitness for that game actor pair F(a,g) for the actors associated with that game g.

$$F_g(A) = Var_{(a \in A_{on}(g))} F(a,g)$$

The intuition behind this definition of actor molecule fitness is that we wish to select for actor molecules in the population that are good at games. The intuition behind the fitness of a game molecule is that we wish to select for games that maximize the predicted rate of fitness increase in the actor population. To do so we exploit a simple result from evolutionary biology, Fisher's fundamental theorem of natural selection. The theorem states that 'the rate of increase in fitness of any organism at any time is equal to its genetic variance in fitness at that time'[1] (Fisher 1930). This applies under the following conditions: i. There must be no epistasis, that is, the fitness contributions of alleles in the units under selection must be additive and linear. ii. There must be no linkage disequilibrium, i.e. no non-random association of alleles at two or more loci in the genotype of the molecule under selection. iii. There must be no frequency dependent selection, i.e. the fitness of one actor molecule must be independent of its frequency in the population of actor molecules.

Whilst none of these conditions apply in any realistic and complex system; however, the deviation from the ideal is often sufficiently small that the approximation is valid. Violations of the three conditions above, e.g. epistasis, may cause the theorem to fail in predicting long term fitness changes, for example neutral evolution may be observed in which the population explores a fitness plateau where all solutions are of effectively equal fitness, until one solution happens to find a step to a higher fitness plateau. If such features dominate the system, then other more sophisticated predictors of evolutionary progress must be used to select for games on which progress can be made. For example, maximizing the fitness entropy of offspring, or applying a Rechenberg type game fitness proportional to the proportion of offspring that are superior to the parent on the game, may be explored (Rechenberg 1994). The key principle is that we are using measures of that predict evolutionary progress to determine the fitness of a game.

Price's equation elaborates Fisher's fundamental theorem by including a term due to transmission bias, i.e. mutation, or environmental change. Fisher's equation corresponds to the first part of the Price equation and states that the rate of change of a trait is proportional to the covariance between fitness and that trait. In short, Fisher and Price reveal that by observing an instantaneous property of the population of actors, a game unit can determine whether progress in that game is expected in the future. Contrary to other approaches (Oudeyer, Kaplan et al. 2007; **Baranes** and **Oudeyer 2009**), we do not explicitly use regression methods to predict the rate of change of fitness of unit based on sampling performance over time, instead, Fisher and Price's insights permit an instantaneous measure of variance to stand in for a rate of change of fitness. Without realizing this fundamental relation to Fisher and natural selection, the strategy of defining the fitness of a population of tests as the variance in another population of models has been proposed by Bongard and Lipson (Bongard and Lipson 2005), however, they consider not the evolution of behaviours, but specifically of forward models or dynamic equations. The algorithm described here is considerably more general than evolving models of the world.

The high-level pseudo-code for the algorithm is outlined below.

1. Initialize population of actor molecules
2. Initialize population of game molecules
3. While True:
4.     While ($f_{aj}$ values not stably approximated):

---
[1] Fisher, R.A. (1941) Ann. Eugen. I I, 53-63

5. Execute a poorly sampled actor molecule $j$
6. Update $f_{aj}$ over all valid N games for that actor molecule
7. Update $f_{gi}$ based on stable $f_{aj}$ values of actor molecules.
8. Replicate and mutate actor molecules
9. Replicate and mutate game molecules
10. Transfer stable actor/game molecules to Long Term Memory

Some critical features of the encoding of actor and game molecules that permit open-ended evolution are now described. Let us first consider actor molecules. Specifically, the molecules are constructed such that they exhibit systematicity and compositionality (Fodor and Pylyshyn 1988), see Part 1 for a full discussion. In other words they are physical symbol systems analogous to atoms and molecules in chemistry. The function of an atom is systematically determined by its structure, and atoms can be composed into molecules whose function is determined by the arrangement and structure of atoms that compose the molecule. However, in contrast to all other cognitive architectures, e.g. CopyCat (Hofstadter and Mitchell 1995), SOAR (Newell 1990), ACT-R (Anderson 2007), and CLARION (Sun 2002), our system integrates connectionist and symbolic systems in a manner that preserves the full power of the physical symbol system as required by Fodor and Pylyshyn; see Gary Marcus' book "The Algebraic Mind" for a detailed criticism of previous attempts at hybrid cognitive architectures within the domain of symbolic connectionism (Marcus 2001). There is no conflict in our framework with other machine learning or connectionist approaches because all the non-symbolic features of a wide range of supervised and unsupervised learning algorithms found in machine learning textbooks can be possessed inside an atom in our architecture.

**Design of Actor Atoms**

Each actor unit or actor 'atom' as we will refer to it, is of the form shown in Figure 2. Each atom contains:

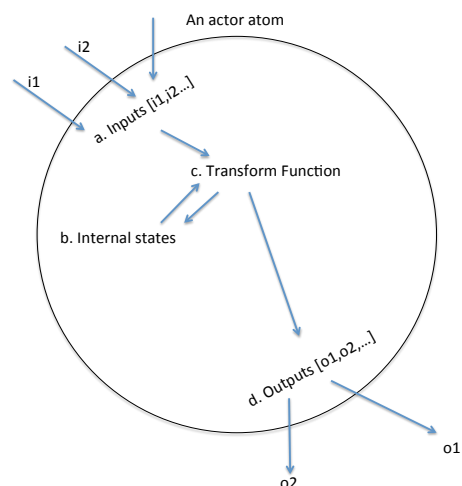

Figure 2. General structure of an actor atom

a. A list of strings $i$ that specify from where the input data will come into the atom, i.e. it labels the registers from which values are taken as input to the

   atom. These registers are found in a working memory that atoms can read from and write to asynchronously[2].
b. A set of internal registers *r* that can store states.
c. A transform function *T* that uses the input to calculate an output. Any transfer function is permitted, e.g. a feed-forward neuronal network, a logical function, a stochastic hill climbing algorithm, covariance matrix adaptation evolution strategy, Q-learning, in fact any function is allowed, and the greater the diversity of such functions the better. In later implementations users will be able to submit such atoms to an online web interface.
d. A list of strings *o* that specifies to where in working memory the output data will be written when it is produced by c.
e. A list of fitness *f* values obtained from game molecules in the game population.

Thus, an atom is specified by a tuple $\{i,r,T,o,f\}$. The data inside an atom is encapsulated and not directly accessible by other atoms. Atoms send information to each other via the registers in working memory. This information in a register can for example be as simple as a signal for other atoms that observe this register to turn on or off, or as complex as a Turing complete computer program. However, in the examples provided, the most complex information passing that is needed is to pass a short list of floating point numbers.

**Binding of Actor Atoms to form Actor Molecules**

The functional binding or linking of atoms to each other is via interactions mediated by working memory. Working memory consists of a dictionary of key-value pairs, where the keys are the entries in the input and output tables of atoms that refer to the registers in working memory that contain the values. It is the values that are effectively passed between atoms. The linkage of atoms in this way creates actor molecules, which are defined as directed graphs of information flow between active atoms. These actor molecules are embedded within a larger network of potentially overlapping information flows between atoms.

Sensory and motor grounding is achieved in the following way. There is a list of dictionary entries to which raw sensory data is written, and from which values are read to control raw motor output. Actor atoms can read from and write to these special registers.

**Design of Game Atoms**

Game atoms define the fitness of actor atoms. A game atom is defined by the following elements:

a. A list of keys *i* in working memory from which input is obtained
b. A function specification *T* for operating on these inputs to produce a fitness value to be assigned to the actor molecule and its atoms.

---

[2] The ALMemory structure of the Naoqi API for the NAO robot produced by Aldebaran Robotics is used for this working memory structure. It is simple to use and thread safe.

    c. A fitness-value *f* that stores the variance of fitness components in action atoms obtained on this game

An example of a game atom is one that observes a working memory location written to by an actor atom and gives fitness to that actor molecule in proportion to the accumulated value at that location in working memory over one trial. More complex functions of the value may define fitness, or fitness may be a function of many locations in working memory.

**Binding of Game Atoms to make Game Molecules**

Game atoms can be bound to each other in the same way as actor atoms in the following manner. An atomic game specifies the actor molecule fitness directly. This value is passed for storage in the relevant actor molecule, however, a molecular game can contain game atoms that are written to memory itself, and this may result in activation and information passing to other atoms in the game population which themselves calculate further functions on the output of upstream game atoms. In this way, more complex games can be specified e.g. minimize the horizontal distance of the visual image of the tip of a finger and the tip of another finger. This may be a game that the robot can have in mind when assigning fitness to its actor molecules.

**Initialization of Actor and Game Population**

The actor population is initialized using innately specified behaviors loosely corresponding to reflexes. Some examples of such reflexes are given later. These reflexes may be tagged such that they can never be completely erased from the population of actions.

The game population is initialized with a set of innate games that are designed to complement the set of innate 'reflexes'. For example, if a reflex actor molecule exists that moves the head to foveate on a face, then it may come with a matching game molecule that assigns fitness to it in proportion to the speed at which foveation was achieved over a trial.

**Actor Molecule Fitness Assessment Loop**

1. While ($f_{aj}$ values not stably approximated):
2.    Execute a poorly sampled actor molecule *j*
3.    Update $f_{aj}$ over all valid N games for that actor molecule
4. Update $f_{gi}$ based on stable $f_{aj}$ values of actor molecule.

The $f_{aj}$ values for actor molecules must be approximated and this is the primary concern of the robot's action selection method to be described below which chooses which actor atoms to activate preferentially. Typically, the actor atom that has been activated the least number of times is activated. Once the fitness of actor molecules has been stably approximated, or typically after a fixed number of evaluations per atom, the fitness of the game molecules can be calculated.

**Action Selection Method**

At the beginning of the while loop above, all action atoms are inactive. The action selection algorithm's task is to execute a poorly sampled actor unit and run the consequences of starting the network from that atom, in order to obtain stable fitness scores for all the actor molecules in the network so that selection can take place fairly. It does this by choosing between the actor atoms that have only raw sensory inputs (and no inputs from other atoms), and activating that atom which has been activated least often in this generation.

Once an initiator atom has been activated by the above procedure the process of downstream actor atom activation is started through the currently active atoms writing activation signals that are read by other atoms from working memory. Typically embedded within the atom function there is a condition statement that says whether the atom will become activated as a function of its inputs. This is typically specified as a logical function of its input states. The set of active actor atoms in a trial constitutes the actor molecule. Thus a molecule of actor atoms is activated over the trial that runs for a fixed maximum period of time, or until a special termination atom is activated. Some examples of such processes are shown later in the paper.

At each time step during execution of an actor molecule, game molecules check working memory and send their output to update the corresponding states in the actor molecules that are responsible for those states in working memory. These actor atoms store the fitness component due to that game moleucle in their state registers. These scores are accumulated within the actor atom over one game execution, and a vector of such accumulated scores are obtained for each trial in which that atom participates over that generation. It is this vector that is used to determine whether a stable approximation of fitness has been obtained.

**Game Atom Fitness Assessment**

Once a stable approximation of actor atom fitness has been obtained, the loop is exited and game atom fitness is updated as a function of the fitness scores accumulated in the actor atom population. The fitness function for a game molecule is the variance of fitness components of actors on that game.

**Replication and Mutation of Actor Atoms**

The replication of actor atoms is of two types that occur with different frequencies *a* and *m*. Type A mutations are atomic mutations which involve an actor making a copy of itself in the following ways, see Figure 3a.

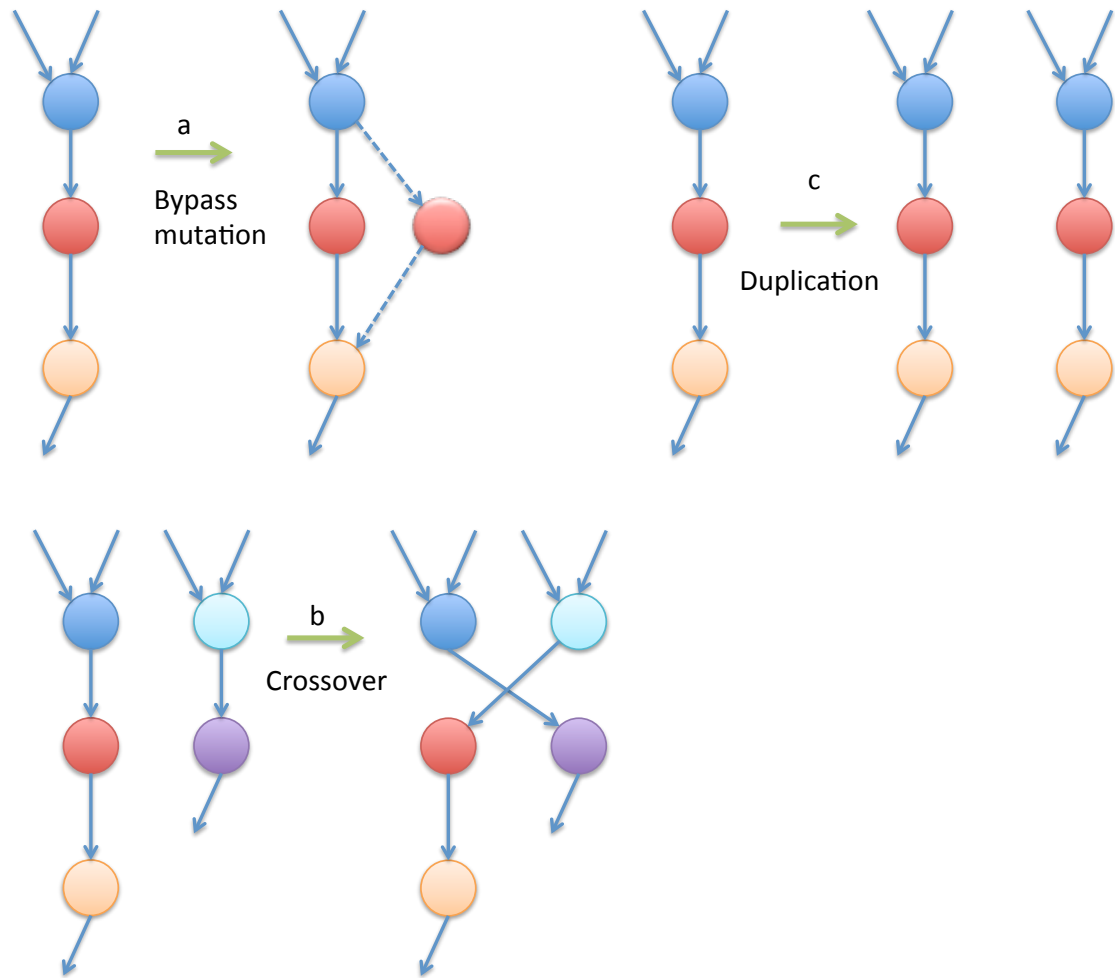

Figure 3. Mutation, crossover, and duplication operators on behavior atoms.

The entire tuple that describes the atom, i.e. the states, functions, inputs and outputs of the actor atom are copied to the offspring actor atom. This results in a bypass route of activation coming into existence in an actor molecule, but where the offspring atom is still connected to the original molecule. Mutation during the copy operation is of two types; generic and specific. Generic mutations are those that apply to all atoms and they consist of homogenous mutation in input space, i.e. the identity and number of inputs to the atom, and in output space, i.e. the identity and number of outputs of an atom to working memory. A crossover operation is another type A operation, see Figure 3b.

Specific mutations are those explicitly encoded within the atom function itself, and that are encoded at initialization of the atom. For example, if the atom function is a neural network then the mutation operator may be Gaussian variation to each of the weights in the neural network. If the atom function is expressed as a mathematical equation, then genetic programming type operators acting on the form of the equation may be more suitable.

Type M mutations are molecular level variation operators rather than just atomic level operators. The canonical example being a complete duplication and divergence event in which an entire molecule active in one game is copied to a new disconnected location in the actor network, see Figure 3c.

There are several subtleties involved in increasing the probability that a mutant is functional. When an atom is copied there arise two possible routes of activation through the molecule. Which one is taken depends on competition between the two atoms in the same molecule. For example, one atom may calculate the Euclidean distance between inputs and the other atom may calculate the Manhattan distance between inputs. Competition between the atoms in a molecule takes place by the path evolution algorithm (Fernando, Vasas et al. 2011). The strength of alternative outflow paths to competing atoms is adjusted in proportion to the reward obtained by the downstream atom using Oja's rule. With entire molecular duplications the path evolution methods are not needed.

Note, that when an actor atom is replicated the copy writes to a new memory location, and so game atoms that observe the memory location of the parent actor atom must also be modified to observe the memory location of the offspring actor atom.

**Replication and Mutation of Game Atoms**

The operations on game atoms are identical to those on actor atoms. Consider an example of a game atom that defines a game in which the x,y,z coordinates of the center of gravity (CoG) of the body [a value that is available in the ALMemory of the NAO robot, but not in human infants] must be changed as much as possible within 10 seconds. When an action molecule is executed, its fitness on this game can be calculated. Random variants to this game generate quasi-species of adjacent games, for example.

1. Minimize the distance moved of the CofG.
2. Maximize distance moved of CoG while keeping a face in the visual field
3. Maximize distance moved of CoG while minimizing energy required to move.
4. Maximize distance moved in retinal coordinates of a visual object.
5. Maximize distance moved in joint angle space of a subset of joints.

These games all share atomic components and can be generated by M and A type mutation operators. Which of these games survive and outcompete the others is determined by how much variance there is on that game in the actor population.

**Transfer of Stable Actor and Game Molecules to Long Term Memory**

Combining diversity maintenance and the accumulation of adaptation is the evolutionary computation version of the stability plasticity dilemma (Fernando 2010). The solution used here is to take actor molecules or game molecules that have reached fixation in the population out of the population and into a long-term memory store. There they can be activated in the same way as when they are in the population but they are immutable. In addition, a molecule similarity function is implemented which punishes molecules in the population in proportion to their similarity to molecules already in long-term memory. Over time, adaptive games and solutions accumulate in LTM, permitting the evolving populations to explore new games and solutions. Limiting the kinds of exploration that the agent engages in.

**Games, Solutions, and Adjacent Possible Behaviors**

Given the semi-formal operations defined above, it is now possible to hand-design a set of behavioral molecules and understand the behavior that they will produce in the robot that they control. It is also possible to apply the A and M type variation operators to the specifications and see what kinds of adjacent possible behaviors result from random variation.

The algorithm is implemented in Python using the Naoqi API provided by Aldebaran robotics. This provides an ideal asynchronous modular toolbox that already contains a vast amount of high-level functionality. In effect, each python module or box in the choregraphe framework can be incorporated into an actor atom as defined above. For example, there may be a face recognition atom that takes visual input and outputs the presence or absence of a face in the current visual field, and the string value associated with that face, a sound localization atom may return the angular displacement of the head that is required to rotate the head towards the predicted location of the sound. Such high-level actor atoms can also potentially be submitted by web-users. Conversely, low-level actor atoms also exist. Such a low-level actor atom may for example contain a simple 3 layer feed-forward neuronal network that maps directly from sensory input to motor output, thus implementing a pure reflex action. Atoms may contain any transfer function, e.g. a liquid state machine, an SVM etc, etc. In addition, atoms can contain unsupervised learning algorithms, e.g. EM algorithms, K-means clustering, Principle Component Analysis, Independent Component Analysis, mutual or predictive Information calculations, and these may be parameterized in different ways, with variation operators defined on the functions. Thus, the network of controllers that is produced is entirely general. Of-course, the more complex the atom the more brittle it will be to mutation. The research program generated by this paper is to discover the evolvable grammar of action used by a brain. Let us now consider some specific action molecules and their associated game molecules.

*Reacting to Resistance*

The action molecule described is capable of controlling the robot to hill climb in elbow joint angle space in order to maximize the prediction error of the angle of the elbow, see Figure 4. In other words, it tries to do actions that result in unpredictable elbow positions, for example, if an obstruction is encountered then it will explore those joint angles containing the obstruction preferentially because these are the joint positions that the forward model does not currently get right. Four atoms encode the closed-loop behavior. The first atom takes the elbow angle and motor command as input, and uses these to update a forward model of the elbow angle. This requires storage of elbow angle and motor command for one time-step within the atom. The output that is written to working memory is the predicted elbow angle at the next time-step.

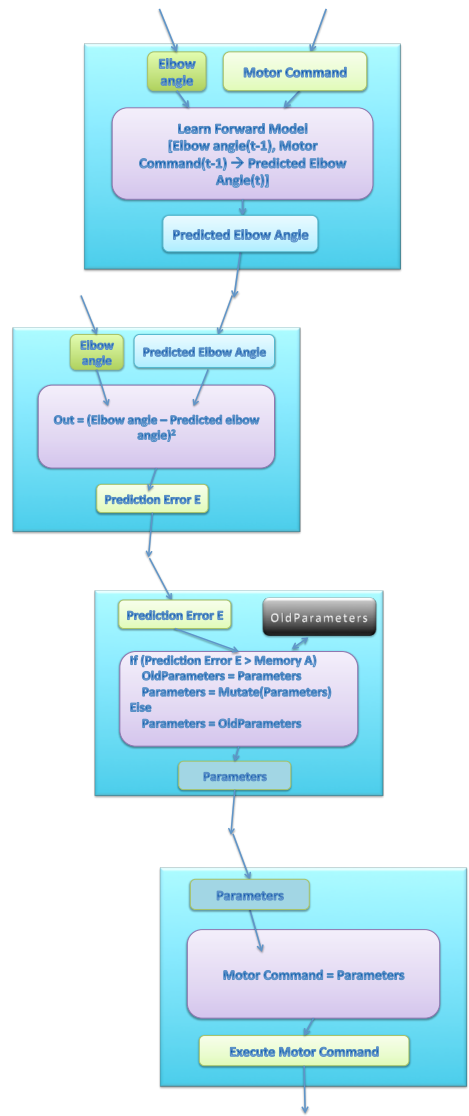

Figure 4. A 4-atom action molecule for optimal elbow exploration, with the property that it 'enjoys' exploring elbow joint positions associated with unexpected angle effects of motor commands to the elbow joint, i.e. reacting to resistance.

The second atom is then activated. This takes the predicted elbow angle and the actual elbow angle and calculates the squared error between these two inputs and writes to memory the output that is the prediction error of elbow angle. This atom then activates the third atom. The third atom in this chain molecule is a hill-climber atom that takes an input that it always treats as reward (in this case the input is prediction error of the elbow joint model. It tries to maximize reward by using hill-climbing on a parameter vector. If reward increases, the system keeps the current parameters and does further exploration, else the system reverts to the old parameters. Finally, these parameters are written to working memory that activates the final molecule to interpret these parameters as motor commands, which means to write them to the motor registers that are immediately executed. When this motor command to move the elbow is written to memory this reactivates the first atom in the molecule, thus causing the loop to iterate. After a fixed time period the loop is stopped, all atoms being inactivated.

An example of a game associated with the above molecule is an atom that simply requires that the prediction error output of the second atom is maximized over some time period e.g. 10 seconds. Consider mutations of the above behavior molecule. The following one-step mutants are possible.

1. Bypass mutation of atom 1: Get input from shoulder angle not elbow angle, or from foot motor not elbow motor.
2. Bypass mutation of atom 2: Mutate transformation in atom 2, so that prediction error = Elbow angle – Predicted elbow angle, not the square of the later.
3. Bypass mutation of atom 3: Modify parameters of S.H.C. e.g. probability of accepting a worse set of parameters. Or try to minimize rather than maximize the input.
4. Bypass mutation of atom 4: Control another motor, e.g. the foot or the neck.

Many of the mutants will result in molecules that are non-functional, e.g. stochastic hill climbing will not be able to hill climb if one is trying to control the elbow angle by moving the foot. Molecules of this type will score poorly on the above game and will be unlikely to be selected on the basis of performance on that game. However, mutant 3 may result in interesting behavior where the agent tries to minimize not maximize prediction error. This would result again in poor performance on the original game however. For this variant to survive, a new game mutant is required that rewards minimization of the output of atom 2.

*Orienting to Objects*

Visual orienting to objects can be undertaken by a range of possible behavior molecules. Figure 5 shows one possible molecule. The first atom receives the raw sensory input and uses high level visual processing routines to extract a vector of x,y retinotopic coordinates of salient visual objects. For example the object recognition module in Naoqi by Aldebaran can be used. This output is read from working memory by an inverse model atom. This atom has learned a model mapping the desired visual location of the object to be foviated, and the current neck angle, to the desired neck angle required to achieve foviation. This output is then sent to a standard motor execution atom that takes the input, contains a list of motors to be commanded and writes to the motor register to execute the action.

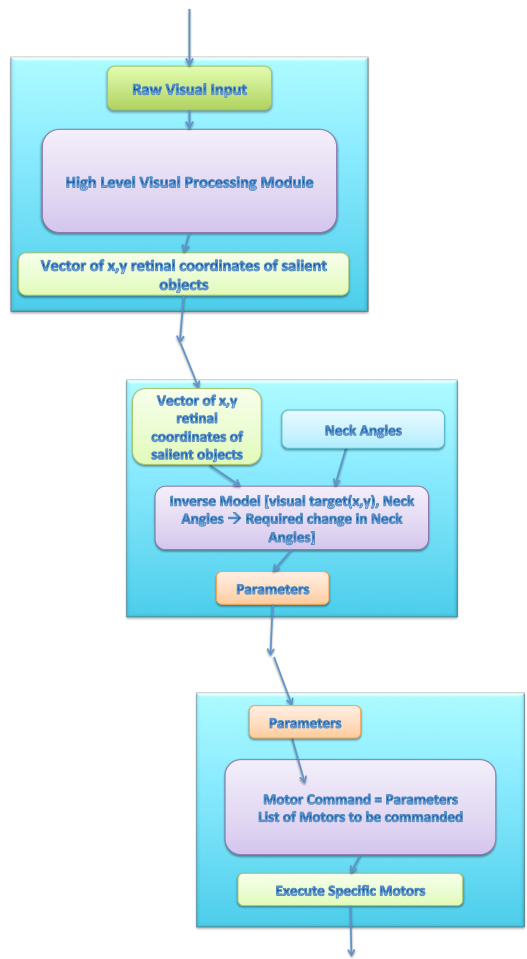

Figure 5. Behaviour molecule for orienting to a salient visual input by moving the neck.

A game that stabilizes the above molecule would be to minimize the distance between salient visual objects. Variants of the 'orienting to objects' behavior that can be produced by mutation of the above molecule include,

1. Turning away from objects, achieved by replacing the inverse model with another model, and by mutating the game so as to maximize the visual distance between salient objects.
2. Orienting to different kinds of salient visual object, achieved by modifying the visual processing module in atom 1.
3. Moving the object in a specific way over the visual field, e.g. oscillating it in the visual field, achieved by modifying the inverse model in atom 2, and by modifying the game so as to reward high variance over time in the visual distance between salient objects.

*Visual Reaching*

Figure 6 shows a molecule that contains a bypass, i.e. two competing routes of activation that compete for control of a limb. The two routes compete for control and may be strengthened and weakened by the path evolution algorithm over the course of development. Visual reaching could be learned and remembered using a high-level processing atom that generates an output when the hand is visible as well as another salient object (which may be another hand). The x,y coordinates of the salient objects are written to working memory as before. A range of atoms may read these as input, for example an inverse model atom may take these values and control the muscles of the left arm to reach the object. Another atom may implement stochastic hill climbing on the joints of the left arm with the inverse of the distance between the hand and the visual object as reward to be maximized. This later chain may be more effective at reaching when the inverse model has not been properly learned. In this way, one path through the system can maintain inefficient functionality while another path learns a more efficient controller.

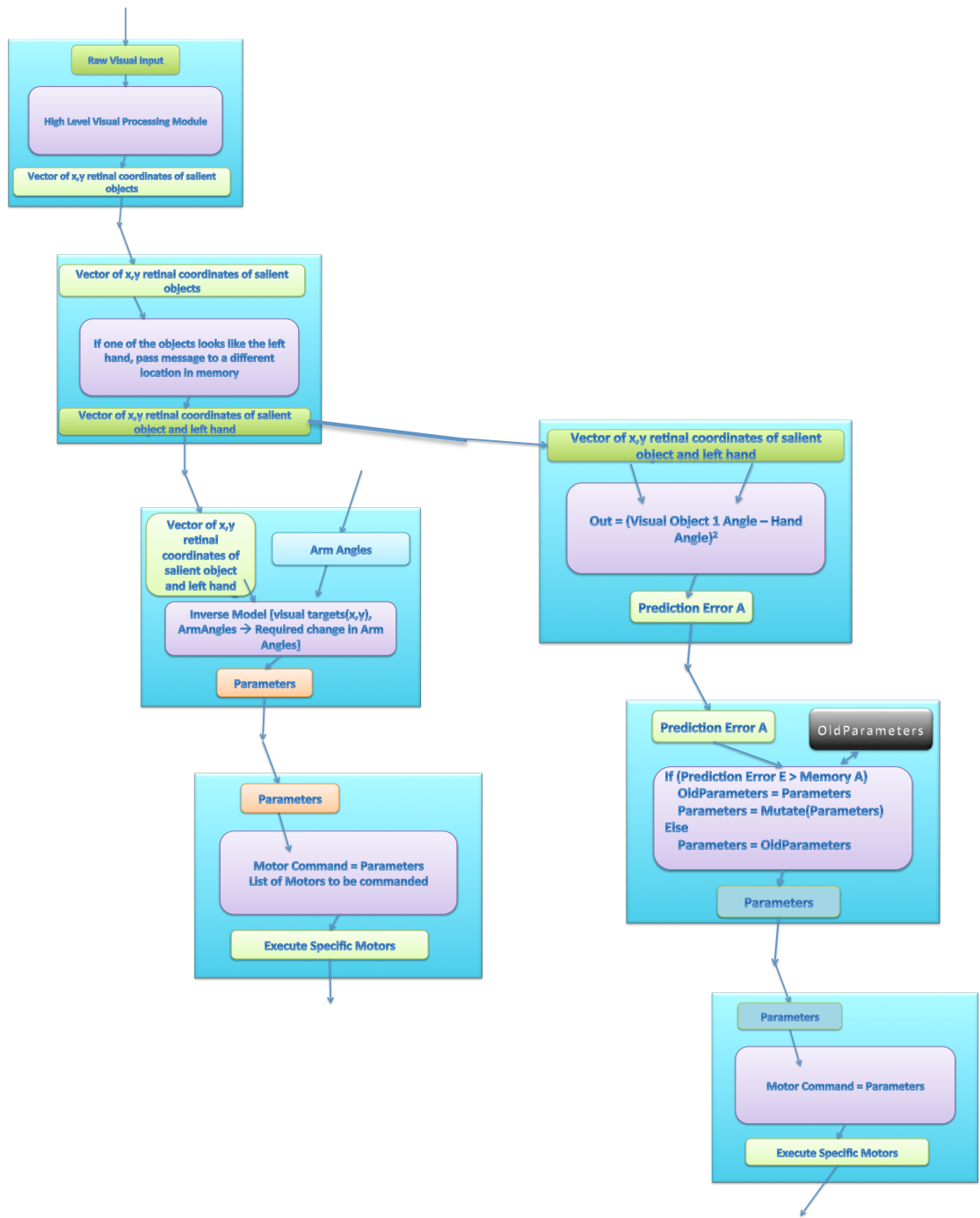

Living Machines 2013 Natural History Museum, London.

Figure 6. Two pathways for visual reaching with the left arm. The one on the left uses an inverse model and the one on the right uses stochastic hill climbing. Both aim to do the same thing, visual reaching, in different ways.

*Maximizing Mutual Information between Effectors and Sensors*

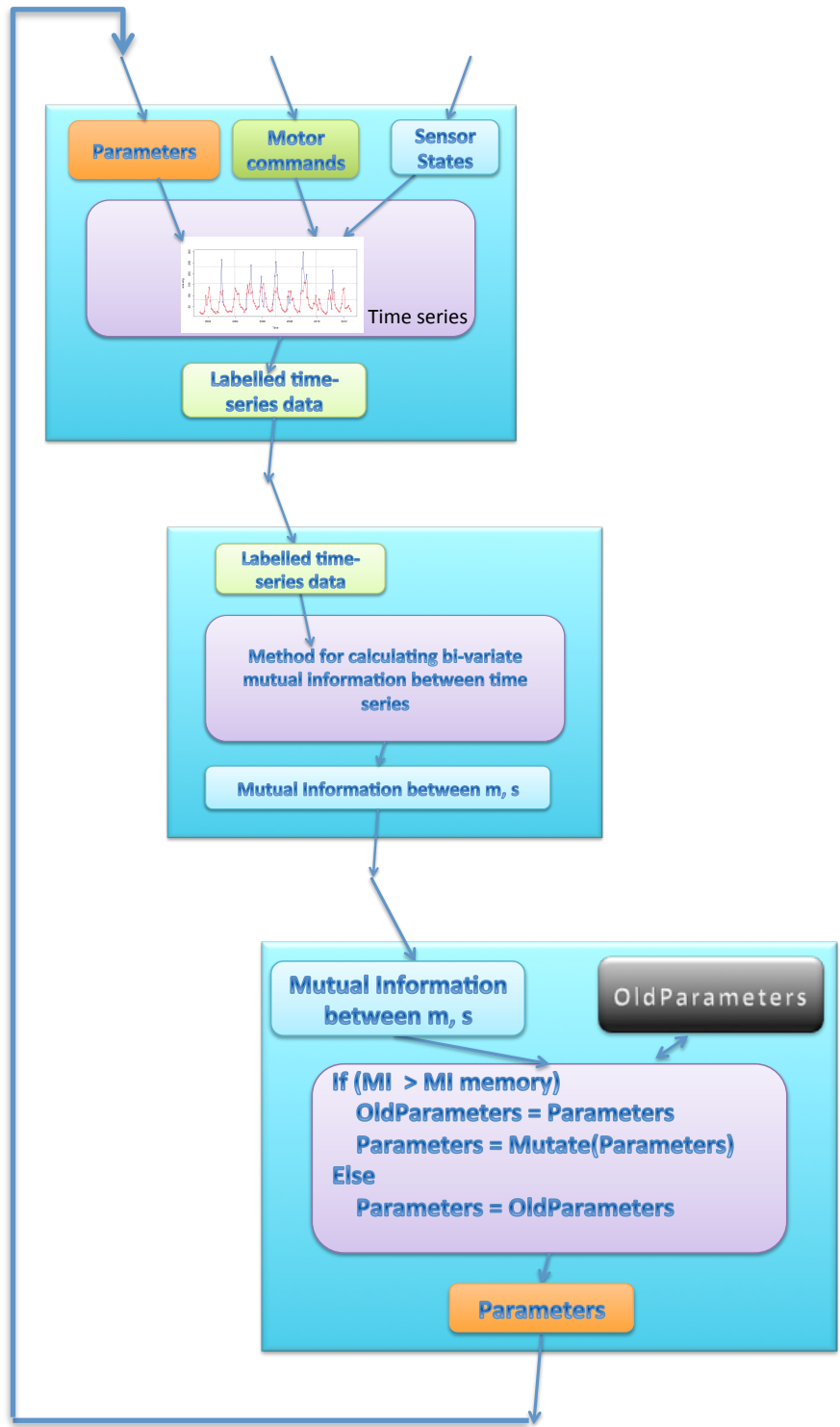

Figure 7. Stochastic hill climbing in the space of sensor-motor pairs to find those with high mutual information between them.

Figure 7 is an example of an unsupervised learning molecule that acts in order to detect sensorimotor contingencies, and stores combinations of effector-sensor pairs that have high mutual information. These pairs can be used by other systems to structure exploration at a later stage. This molecule detects one such sm contingency with the highest mutual information between the sensor and motor time windows. Notice, there is no motor control in this molecule, it does not influence behavior at all, only observes behavior. In a sense, it is a molecule of thought, not of action. It is ofcourse possible to have atoms with somewhat different information measures, for example predictive information may be used, or Granger causality. The cognitive architecture presented here is Catholic, an atom will be preserved if it is stable within the co-evolutionary dynamics of actions/thoughts and games, whatever that atom contains.

**Specific Examples of Intrinsic Motivation Games**

In the field of intrinsic motivation and artificial curiosity and creativity, an entire quasi-species of functions have been proposed which an agent may want to maximize. Our architecture permits not only the representation of this quasi-species of ideas, but the transition between them using mutations to the game specifications. Without exception, all these functions can be encoded within the action and game description language described here. For example, Schmidhuber's idea of maximizing the first derivative of compressibility can be encoded in a game molecule that contains a recorded time series, and a range of compressors (Schmidhuber 2006). Similarly Oudayer's idea of maximizing the first derivative of predictability or explicitly sampled competence progress over time can also be encoded by a game molecule that records time series and calculates compression progress on them (Oudeyer, Kaplan et al. 2007).

In all cases, the overall survival of a game is determined by the high-level fitness function for game molecules. This is simply fitness variance of actors on that game. For example, if Schmidhuber's game results in a high variance of actors in their ability to compress data, then Schmidhuber's game may well survive. If Oudayer's game results in a high variance of competence progress in actors, then Oudayer's game may survive. In fact, the next stage of this research is to implement our algorithm online so that Schmidhuber, Oudayer and others can submit to Darwinian neurodynamics which determines whether Schmidhuber's game is an interesting game to play or not. Schmidhuber would also be encouraged to submit mutation operators so that the adjacent possible of function variants around his original idea of compressibility could be automatically explored by our algorithm.

**Discussion and Conclusions**

It remains to be seen what kinds of co-evolutionary pathology arise from specific implementations of this architecture and what kinds of specific mutation operator are required to produce robust behaviors with non-brittle outcomes. However, we believe that the above architecture defines a workable correct broad outline for open-ended creative cognition. We note that a related and powerful algorithmic approach in the domain of language is the fluid construction grammar that evolves linguistic constructions (Steels, in press). It does a constrained search on structured representations, just as our algorithm does, however, the algorithm presented here is more general.

The current development version of the architecture includes a 3-atom molecule that evolves by M type operators and has one game molecule implemented. All development versions will be available on github[3]. We intend to implement the architecture also in Processing for a range of 'janky' arduino robots all controlled by the same creative algorithm.

**Acknowledgements**

Many thanks to MAT student Mr Christopher Jack for providing the reaching with resistance example. Thanks to Alex Churchill and Mate Varga for reading the manuscript and providing comments. Thanks to Goren Gordon for useful discussions during an IM-CLEVER meeting at FIAS. Thanks to the UK NAO Group, Dave Snowden and Christie Nel for help with the code. Thanks to Emilia Maria Garcia for help with formalization. Finally thanks to Eors Szathmary for guidance and comments.

---

[3] https://github.com/ctf20/DarwinianNeurodynamics/

Living Machines 2013 Natural History Museum, London.